\documentclass[bst/sn-mathphys]{sn-jnl}

\jyear{2021}%

\theoremstyle{thmstyleone}%
\usepackage{amsmath,amsfonts,bm}

\def\eqref#1{equation~\ref{#1}}

\def\1{\bm{1}}

\DeclareMathAlphabet{\mathsfit}{\encodingdefault}{\sfdefault}{m}{sl}
\SetMathAlphabet{\mathsfit}{bold}{\encodingdefault}{\sfdefault}{bx}{n}

\DeclareMathOperator*{\argmax}{arg\,max}

\usepackage{amsfonts} 
\usepackage{array}
\usepackage{amsthm}
\usepackage{mathtools}
\usepackage{enumitem}

\usepackage{bm}

\usepackage{graphicx}
\usepackage{amsmath}
\usepackage{booktabs}
\usepackage{amssymb}
\usepackage{multirow}
\usepackage{multicol}
\usepackage{makecell}
\usepackage{overpic}
\usepackage{bbding}
\usepackage{wasysym}
\usepackage{arydshln}
\usepackage{latexsym}

\theoremstyle{thmstyletwo}%

\theoremstyle{thmstylethree}%

\begin{document}

\title[Tuning Synaptic Connections instead of Weights by GA in SPN]{Tuning Synaptic Connections instead of \\ Weights by Genetic Algorithm in \\Spiking Policy Network}


\author[1,2]{\fnm{Duzhen} \sur{Zhang}}

\author[1,2]{\fnm{Tielin} \sur{Zhang}}
\equalcont{Corresponding author}

\author[1,2]{\fnm{Shuncheng} \sur{Jia}}

\author[1,2]{\fnm{Qingyu} \sur{Wang}}

\author[1,2,3]{\fnm{Bo} \sur{Xu}}
\equalcont{Corresponding author}

\affil[1]{\orgdiv{School of Artificial Intelligence}, \orgname{University of Chinese Academy of Sciences}, \orgaddress{\city{Beijing} \postcode{100049},  \country{China}}}

\affil[2]{\orgname{Institute of Automation, Chinese Academy of Sciences (CAS)}, \orgaddress{\city{Beijing} \postcode{100190},  \country{China}}}

\affil[3]{\orgname{Center for Excellence in Brain Science and Intelligence Technology, CAS}, \orgaddress{\city{Shanghai} \postcode{200031},  \country{China}}}




 \abstract{Learning from interaction is the primary way that biological agents acquire knowledge about their environment and themselves. Modern deep reinforcement learning (DRL) explores a computational approach to learning from interaction and has made significant progress in solving various tasks. However, despite its power, DRL still falls short of biological agents in terms of energy efficiency. Although the underlying mechanisms are not fully understood, we believe that the integration of spiking communication between neurons and biologically-plausible synaptic plasticity plays a prominent role in achieving greater energy efficiency.
Following this biological intuition, we optimized a spiking policy network (SPN) using a genetic algorithm as an energy-efficient alternative to DRL. Our SPN mimics the sensorimotor neuron pathway of insects and communicates through event-based spikes. Inspired by biological research showing that the brain forms memories by creating new synaptic connections and rewiring these connections based on new experiences, we tuned the synaptic connections instead of weights in the SPN to solve given tasks. Experimental results on several robotic control tasks demonstrate that our method can achieve the same level of performance as mainstream DRL methods while exhibiting significantly higher energy efficiency.\footnote{The source code, saved checkpoints and results are available at~\url{https://github.com/BladeDancer957/SPN-GA}.}}




\maketitle

\section{Introduction}\label{sec1}

When we consider the nature of learning, the first thing that comes to mind is the idea of biological agents learning by interacting with their environment. Reinforcement learning (RL) is an algorithm that attempts to study this biological mechanism of how agents learn from their interactions, from a computational perspective \cite{sutton2018reinforcement}. In RL, an artificial agent interacts with the environment in a trial-and-error manner and learns an optimal policy by maximizing the accumulated rewards. In modern deep reinforcement learning (DRL), agents controlled by deep neural networks (DNNs) and optimized through gradient-based algorithms have proven to be capable of solving various tasks, ranging from video games \cite{mnih2015human, vinyals2019grandmaster} to robotic control \cite{duan2016benchmarking,lillicrap2016continuous,zhao2023ode}.


Although modern DRL has made remarkable progress, it still faces a challenge of high energy consumption. This includes both high inference and optimization energy consumption, which has been a major concern in real-world applications, especially for robot-control tasks that have limited onboard energy resources \cite{tang2021deep}. Specifically, the high inference energy consumption is due to the dense communication of continuous floating-point values, which is required by a deep policy network (DPN), a DNNs-based policy network, resulting in high computational costs.
Moreover, the high optimization energy consumption mainly arises from performing backpropagation (BP) \cite{rumelhart1986learning}. For instance, the computational cost of layer-by-layer BP for calculating gradients in gradient-based algorithms is significantly high, particularly when optimizing the parameters in a DPN using stochastic gradient descent/ascent.


In summary, despite its power, DRL is still far from being energy-efficient compared to the learning mechanism of biological agents. During learning from environmental interaction, biological agents typically exhibit ultra-low energy consumption \cite{salem2018biological, blanchini2011structurally, whitacre2012biological}. Recent studies have shown that the biological brain can perform complex calculations with only 20 watts of energy \cite{attwell2001energy, howarth2012updated}, while existing computing systems often require at least an order of magnitude higher energy consumption to achieve the same task \cite{cox2014neural}.
Although it is not yet fully understood how biological agents learn so efficiently, we believe that the integration of spiking communication between neurons and biologically-plausible synaptic plasticity plays a fundamental role.


Drawing on biological intuition, we propose an energy-efficient alternative to DRL by optimizing a spiking policy network (SPN) using a gradient-free genetic algorithm (GA). Our SPN, which is used to infer actions from state observations, is inspired by the biological sensorimotor neuron pathway found in insects \cite{bidaye2018six, chen2022deep}. The neurons in the SPN communicate via event-based binary spikes, which can reduce inference energy consumption.
To reduce the optimization energy consumption, we design a gradient-free optimization paradigm based on GA, which directly conducts a SPN search. Furthermore, we highlight the importance of synaptic connection tuning as opposed to solely tuning synaptic weights, which is the traditional approach to optimizing a policy network.
Inspired by biological research, where the brain forms memories by forming new synaptic connections and rewiring existing ones based on new experiences \cite{black1990learning, dayan2011neuroplasticity, kleim2002motor}, we only tune the synaptic connections of the SPN to solve specific tasks, without adjusting the synaptic weights.


    

Our main contributions can be summarized as follows:

\begin{itemize}
\item Conventional DRL algorithms are computationally expensive and become more so as task complexity increases. To address this issue, we construct a biology-inspired SPN optimized by a gradient-free GA, providing an energy-efficient alternative to DRL. Our approach is innovative and significant for energy-efficient computation in the RL field, where the GA algorithm can replace big memory buffers to achieve a similar independent identically distributed distribution of training data.

\item We adopt the viewpoint of direct synaptic connection tuning instead of weight tuning by GA in SPN to solve RL tasks. Our surprising finding is that modifying synaptic connections is not only more efficient but also more fitting for the SPN, where discrete instead of continuous signals are used for information representation.

\item We present extensive experimental results on several robotic control tasks, demonstrating that our alternative method can achieve the same performance level as mainstream DRL methods while exhibiting significantly higher energy efficiency (saving about 1.1$\sim$1.3 times inference energy consumption and 6.5$\sim$87.9 times optimization energy consumption under various tasks). Further analysis verifies the superiority of connection tuning over weight tuning and the necessity of using GA to optimize the SPN.
\end{itemize}

\section{Related Work}

\subsection{RL}

Conventional RL algorithms excel at handling simple, tabular state spaces~\cite{sutton2018reinforcement}. However, they face challenges when dealing with complex state spaces. To address this issue, modern DRL relies on powerful function approximators, such as DNNs, to learn a mapping function directly from the raw sensory state space to the action space. Two main families of gradient-based algorithms are commonly used in DRL: the Q-learning algorithm~\cite{watkins1992q}, represented by DQN~\cite{mnih2015human}, and the policy gradient algorithm~\cite{sehnke2010parameter}, represented by PPO~\cite{schulman2017proximal}.


DQN~\cite{mnih2015human} leverages DNNs to approximate the optimal Q function, resulting in a policy that selects the action that maximizes the Q-value for a given state. Two crucial techniques employed in DQN are the experience replay and the target network. The former stores the agent's experiences into a replay buffer, allowing them to be reused in many weight updates for greater data efficiency. The latter generates the targets in the Q-learning update and addresses the instability problem that arises when combining DNNs and Q-learning. DQN and its improved variants, such as Double DQN~\cite{van2016deep} and Dueling DQN~\cite{wang2016dueling}, are currently advancing towards achieving human-level control in Atari video games~\cite{bellemare2013arcade} with discrete action spaces.


However, the value-based DRL algorithms discussed above have limited processing capacity for continuous action spaces, making them unsuitable for certain scenarios. To address this issue, the PG algorithm~\cite{sehnke2010parameter} was proposed to directly learn the parameters in a DPN. However, it still suffers from the high variance problem, partly due to the probabilistic output for taking actions. To address this, the actor-critic (AC) framework was introduced, where a deep actor/policy network infers an action from a given state, and a deep critic/value network estimates the associated state-value or action-value to guide the actor during optimization. Several improved AC-based PG algorithms have been proposed to improve exploration and data efficiency, including A3C~\cite{mnih2016asynchronous}, TRPO~\cite{schulman2015trust}, PPO~\cite{schulman2017proximal}, DDPG~\cite{lillicrap2016continuous}, TD3~\cite{fujimoto2018addressing}, and SAC~\cite{haarnoja2018soft}. Among these algorithms, PPO is a more general and easier-to-implement algorithm that can be scaled to almost all task scenarios, including video games with discrete action spaces and robotic control with both discrete and continuous action spaces.


Despite the remarkable progress made by DRL algorithms described above, their solutions are often energy-intensive due to the reliance of deep policy networks on dense communication of continuous values and the gradient-based optimization paradigm. To address this issue, we propose a biologically-inspired SPN, optimized by a gradient-free genetic algorithm, as an energy-efficient alternative to DRL.

\subsection{Integrating Spiking Neural Networks (SNNs) with RL}


Recently, there has been a growing body of literature exploring the integration of SNNs~\cite{maass1997networks,DBLP:conf/ijcai/ZhangZJW022} with RL. Early approaches~\cite{o2013spiking,yuan2019reinforcement,mahadevuni2017navigating,doya2000reinforcement,fremaux2013reinforcement} were typically based on local plasticity rules, such as reward-modulated spike-timing-dependent plasticity~\cite{florian2007reinforcement,fremaux2016neuromodulated}. While they perform well in simple tasks such as logic gate functions and random walk problems~\cite{yuan2019reinforcement}, they fail to tackle more complex tasks such as video games or robotic control due to limited optimization capabilities.


To address this limitation, some methods have incorporated SNNs into gradient-based DRL optimization algorithms~\cite{bellec2020solution}. Using the Q-learning algorithm as a base, some approaches directly train deep spiking Q-networks~\cite{chen2022deep,kim2021chip,akl2021porting,liu2021human} or convert trained DQNs into SNNs~\cite{patel2019improved,tan2021strategy} to achieve competitive scores on Atari video games with discrete action spaces. Other methods have proposed a hybrid learning framework tested on robotic control tasks with continuous action spaces. This framework involves training a population-coded spiking actor network in conjunction with deep critic networks using AC-based PG algorithms such as PPO~\cite{tang2020reinforcement,tang2021deep,zhang2021population,DBLP:conf/aaai/ZhangZJ022}. While these methods obtain a SNN-based policy network for inference to solve the high energy consumption of DPN inference, they still rely on gradient-based optimization algorithms, with the high optimization energy consumption overlooked.

\subsection{Neuroscience} 


In neuroscience, a connectome is a comprehensive ``wiring diagram" that maps all neural connections in the brain~\cite{seung2012connectome}. Currently, connectomes for simple organisms like fruit flies~\cite{eichler2017complete,takemura2017connectome} and roundworms~\cite{varshney2011structural,white1986structure} have been constructed. The study of connectomes is motivated by the potential to guide future research on how the brain learns and forms memories through its connections~\cite{anwar2020evolving}.
It is clear that humans, as biological agents, form memories and acquire new skills by creating new synaptic connections, particularly during early childhood~\cite{huttenlocher1990morphometric,tierney2009brain}. The brain then rewires these connections based on new experiences~\cite{black1990learning,dayan2011neuroplasticity,kleim2002motor,bruer1999neural}.
Drawing inspiration from this biological research, we integrate biologically plausible synaptic connection tuning with biologically inspired SPNs to achieve energy-efficient robotic control.

\section{Method}
\subsection{Constructing a SPN}

In RL tasks, an artificial agent interacts with its environment by taking a series of actions based on its state observations and receiving rewards. At each time step $t$, the agent chooses an action $\bm{a}^t$ from a policy $\pi$ based on the current state observation $\bm{o}^t$. The agent then receives a scalar reward $r^{t+1}$ and a new state observation $\bm{o}^{t+1}$, which creates an episode $\epsilon={\bm{o}^0,\bm{a}^0,r^1,...,\bm{o}^{T-1},\bm{a}^{T-1},r^T}$, where $T$ is the length of the episode. The return of the episode $\epsilon$ is the sum of rewards: $R^{\epsilon}=\sum_{i=1}^Tr^t$. To learn a mapping from state observations to actions, we construct a SPN that represents the policy $\pi$, which infers actions $\bm{a}^t$ from observations $\bm{o}^t$.


Our SPN is inspired by the sensorimotor neuron pathway observed in insects, as described in previous research~\cite{bidaye2018six,chen2022deep}. This pathway involves sensory neurons receiving information from the external environment and transmitting it to spiking interneurons that have discrete dynamics. Once the afferent current reaches a threshold, the spiking interneurons generate spikes, which then carry the signal to non-spiking interneurons. The membrane potential of these non-spiking interneurons determines the input current of motor neurons, which are responsible for effective locomotion.

\begin{figure}[tbp]
	\centering  
	\includegraphics[width=1.0\textwidth]{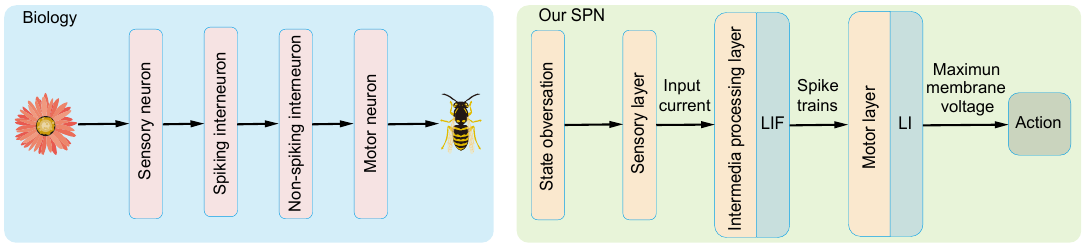}
	\caption{The correspondence diagram between the sensorimotor neuron pathway and our SPN.}
	\label{fig1}
\end{figure}


To align with the biological pathway, our SPN follows a three-layer architecture consisting of the sensory layer, middle processing layer, and motor layer. The correspondence between the sensorimotor neuron pathway and our SPN is depicted in Fig.~\ref{fig1}. For instance, in the context of a robotic control task, the sensory layer acquires a continuous state observation $\bm{o}\in \mathbb{R}^n$ that comprises information such as the angles, velocities, and external forces of each joint of the simulated robot. Here, $n$ represents the dimensionality of the input observation. The sensory information is subsequently transmitted to the middle processing layer, where neurons are modeled using the leaky integrate-and-fire (LIF) \cite{dayan2003theoretical}, which is the most widely used artificial spiking neuron. In contrast, the motor layer neurons are modeled using the leaky integrate (LI) \cite{chen2022deep}, which is a non-spiking neuron similar to LIF but without the capability to fire and reset. We elaborate on the fundamental neuron models employed in our SPN in the subsequent sections.

\paragraph{LIF.} The dynamic firing process at neuron $j$ can be described as:
\begin{equation}
\left\{\begin{array}{l}
\begin{aligned}
\frac{dv_{j,\tau}}{d\tau}&=- g(v_{j,\tau}-v_{rest})+ \sum\limits_{i=1}^n(w_{i,j} \times x_{i,j})o_{i,\tau}\\
v_{j,\tau}&=v_{reset},\ s_{j,\tau}=1\ \ \ if(v_{j,\tau}>v_{th})
\end{aligned}
\label{equa_LIF}
\end{array}\right.
\end{equation}
where $j=1,...,h$, $h$ is the number of neurons in the middle layer, $\tau=1,...,T'$, $\tau$ is a simulation time step, $T'$ is the number of simulation time step (i.e., time window), meaning the same observation will be fed into SPN for $T'$ times, {$v_{j,\tau}$ is the membrane potential of neuron $j$ at time $\tau$ in the middle layer}, $g$ is the decay factor, $v_{rest}$ denotes the rest membrane potential, $s_{j,\tau}$ means the output spike, $x_{i,j}$ is a binary variable indicating whether the synaptic connection from neuron $i$ to $j$ exists, $o_{i,\tau}$ is the input current from upstream neuron $i$ ($i=1,...,n$), $v_{th}$ is the predefined firing threshold, $v_{reset}$ denotes the predefined reset membrane potential, and $w_{i,j}$ is the synaptic weight between neuron $i$ and $j$. In our method, we only tune synaptic connections instead of weights (See Section~\ref{ga} for details). 




The LIF neuron in equation~(\ref{equa_LIF}) captures the temporal dynamics of membrane potential, emulating the behavior of biological neurons. The threshold firing property of LIF neurons results in sparse spike signals. Furthermore, the computation is event-driven, i.e., it is performed only when a spike arrives. This property can reduce the energy consumption during inference, especially when deployed on neuromorphic chips~\cite{akopyan2015truenorth,davies2018loihi}.

\paragraph{LI.} The LI neuron is a special case of the LIF neuron without the firing and resetting mechanism. When the firing threshold $v_{th}$ of a LIF neuron is set to infinity, it becomes a non-spiking neuron. The dynamic equation in equation~(\ref{equa_LIF}) can be simplified as:
\begin{equation}
\begin{aligned}
\frac{dv_{k,\tau}}{d\tau}&= - g(v_{k,\tau}-v_{rest})+ \sum\limits_{j=1}^h(w_{j,k} \times x_{j,k})s_{j,\tau}
\end{aligned}
\label{equa_LI}
\end{equation}
where $k=1,...,m$ and $m$ is the number of neurons in the motor layer.


The LI neurons act as a spike decoder, converting spike trains from the previous layer into continuous values. The membrane potential of the LI neurons is recorded at each simulation time step $\tau$, resulting in $\bm{v}_{\tau}\in \mathbb{R}^m$. The maximum membrane potential is then used to generate the output action:
\begin{equation}
	\left\{
		\begin{aligned}
		\bm{a} &= \mathop{\max}_{1\leq\tau\leq T'} \bm{v}_{\tau}, \text{continuous action space} \\ 
		a &= \argmax(\mathop{\max}_{1\leq\tau\leq T'} \bm{v}_{\tau}), \text{discrete action space}
	\end{aligned}
	\right.
\end{equation}
where $\bm{a}\in\mathbb{R}^m$ represents the torque applied to each joint of the robot for continuous action space, and $a$ represents one of the predefined actions for discrete action space. The forward inference procedure for our SPN is summarized in Algorithm~\ref{ap2}.

\begin{algorithm*}[tbp]
\caption{The pseudocode of forward propagation in SPN}
\label{ap2}
Synaptic weight matrices $\bm{W}^1$ and $\bm{W}^2$ for middle processing and motor layer (randomly initialized and fixed);

Score matrices corresponding to synaptic connections $\bm{C}^1$ and $\bm{C}^2$ for middle processing and motor layer (tuned by GA);

Initialize the score threshold $s_{th}$;

Obtain the mask of synaptic connection $\bm{X}^1=(\text{sigmoid}(\bm{C}^1)\geq s_{th})$ and $\bm{X}^2=(\text{sigmoid}(\bm{C}^2)\geq s_{th})$ for middle processing and motor layer;

Initialize the potential decay factor $g$ and firing threshold $v_{th}$;

$n$-dimensional state observation, $\bm{o}$;

\FOR{$\tau=1$ to $T'$}

\ \ \ \ $\bm{v}^1_{\tau} = g\bm{v}^1_{\tau-1}(1-\bm{s}^1_{\tau-1})+(\bm{W}^1\times\bm{X}^1)\bm{o}$;

\ \ \ \ $\bm{s}^1_{\tau}=\bm{v}^1_{\tau}>v_{th}$;

\ \ \ \ $\bm{v}^2_{\tau} = g\bm{v}^2_{\tau-1}+(\bm{W}^2\times\bm{X}^2)\bm{s}^1_{\tau}$;

Generate output action, $a$:

\IF{Continuous action space} 

\ \ \ \ $\bm{a} = \mathop{\max}_{1\leq\tau\leq T'} \bm{v}^2_{\tau}$;

\ELSE 

\ \ \ \ $a = \argmax(\mathop{\max}_{1\leq\tau\leq T'} \bm{v}^2_{\tau})$;

\end{algorithm*}

\subsection{Tuning Synaptic Connections by GA}\label{ga}
 
\begin{figure}[tbp]
	\centering  
	\includegraphics[width=1.0\textwidth]{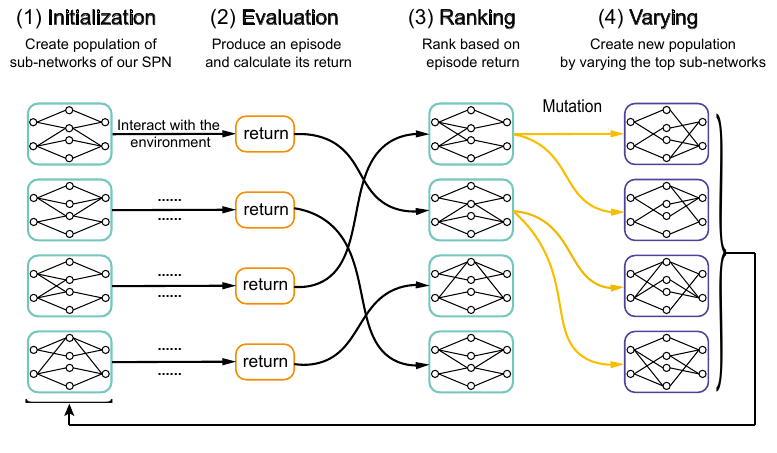}
	\caption{Overview of the evolution of our SPN sub-networks.}
	\label{fig2}
\end{figure}

Our method aims to evolve the sub-networks of the SPN in order to solve a given task for an artificial agent in a simulation environment. During this process, we focus solely on tuning the synaptic connections through GA, rather than placing emphasis on tuning synaptic weights. The evolution of the sub-networks within our SPN can be summarized as shown in Fig.~\ref{fig2}.

\paragraph{Initialization} To begin, we generate the first generation population $\mathcal{P}^0$ consisting of $N$ individuals, each of which represents a vector of SPN parameters $\bm{\theta}^i_j$. This population only contains scores that correspond to synaptic connections without synaptic weights, which are randomly initialized and fixed. In order to reduce variance, we employ mirrored sampling from previous evolution strategy literature~\cite{geweke1988antithetic,brockhoff2010mirrored,salimans2017evolution}. Specifically, we use pairs of perturbations to produce individuals: $\bm{\theta}^0_j = \sigma\bm{\epsilon}\ \text{or}\ \textnormal{-}\sigma\bm{\epsilon}$, where $j=0,...,N\textnormal{-}1$, $\sigma$ represents the standard deviation, and $\bm{\epsilon}\sim \mathcal{N}(\bm{0},\bm{I})$. Once this is complete, we normalize the connection scores between 0 and 1 using the sigmoid function, and then set a score threshold of $s_{th}$. If the normalized score is less than $s_{th}$, we remove the connection, otherwise we keep the connection, resulting in a sub-network. Each SPN sub-network can be considered as an individual.



\paragraph{Evaluation} At each generation $i$ ($i=0,...,G\textnormal{-}1$), where $G$ represents the total number of generations, we evaluate every sub-network to obtain a fitness score. By analyzing the interaction between the sub-network and the environment, we can generate an episode and utilize its return as the fitness score.

\paragraph{Ranking} We rank each sub-network in descending order based on fitness. Then, we perform truncation selection with a ratio of $\eta$, where the top ($\eta*100$)\% of sub-networks will serve as the parents of the next generation. This group of parents is denoted as $\mathcal{P}^i_{\eta}$. To ensure that we obtain the optimal sub-network with greater reliability, we evaluate the top $10$ sub-networks from each generation in $10$ additional episodes (considering these interaction steps as consumed ones during optimization). The sub-network with the highest average return is referred to as the ``elite'' sub-network of the current generation.

\begin{algorithm*}[tbp]
\caption{The optimization process of our GA version}
 \label{ap3}
{\bfseries Input:} number of total generations $G$, population size $N$, standard deviation of perturbations $\sigma$, ratio of truncation selection $\eta$, and fitness score function $F$.

{\bfseries Output:} Elites = [] (store the elite of each generation).

\FOR{$i=0,...,G\textnormal{-}1\ \text{generations}$}

\ \ \ \ Create $i$ generation population of $N$ individuals $\mathcal{P}^i$.

\ \ \ \ \FOR{$j=0,...,N\textnormal{-}1\ \text{individuals}$}
 
 \ \ \ \ \ \ \ \ \IF{$i = 0$}
 
 \ \ \ \ \ \ \ \ \ \ \ {\bfseries Initialization.} $\bm{\theta}^{i=0}_j = \sigma\bm{\epsilon}\ \text{or}\ \textnormal{-}\sigma\bm{\epsilon}\text{,}\ \bm{\epsilon}\sim\mathcal{N}(\bm{0}, \bm{I})$
 
  \ \ \ \ \ \ \ \ \ \ \ \{initialize scores corresponding to all synaptic connections of SPN.\}
 
 \ \ \ \ \ \ \ \ELSE
  
 \ \ \ \ \ \ \ \  \ \ \ {\bfseries Varying.} $\bm{\theta}^{i}_{j}=\bm{\theta}^{i-1}_{j_\eta} + \sigma\bm{\epsilon}\ \text{or}\ \textnormal{-}\sigma\bm{\epsilon}$, $j=0,...,N\textnormal{-}1$, $\bm{\theta}^{i-1}_{j_{\eta}}\in\mathcal{P}^{i-1}_{\eta}$.
 
 \ \ \ \ \ \ \ \ \ \ \ \{Mutation.\}

 \ \ \ \ \ \ \ \ {\bfseries Evaluation.} $F_j = F(\bm{\theta}^{i}_j)$
 
  \ \ \ \ \ \ \ \ \ \{Obtain the sub-network of SPN by sigmoid normalization of connection scores and filtering by score threshold. Based on the interaction between the sub-network and environment, obtain an episode and use its return as the fitness score.\}

 \ \ \ \ \ {\bfseries Ranking.} Sort $\bm{\theta}^{i}_j$ with descending order by $F_j$

 \ \ \ {Perform truncation selection with a ratio of $\eta$ to produce the next generation's parents $\mathcal{P}^i_{\eta}$.}

 \ \ \ \ \ {Set elite candidates for generation $i$ (top 10):}
 
 \ \ \ \ \ \ \ \ {$C^i \leftarrow \bm{\theta}^{i}_{1...10}$.}
 
 \ \ \ \ \ {Select elite for generation $i$:}

 \ \ \ \ \ \ \ \ {$\text{elite}^i \leftarrow \argmax\limits_{\bm{\theta}\in C^i} \frac{1}{10}\sum_{k=1}^{10}{F(\bm{\theta})}$.}
 
 \ \ \ \ \ {Elites\ +=\ [$\text{elite}^i$]}

{\bfseries Return:} Elites
 \end{algorithm*}
 


\paragraph{Varying} Next generation population $\mathcal{P}^{i+1}$ of $N$ sub-networks is created by varying the parents obtained via truncation selection in the previous step. This process of variability is typically referred to as ``Mutation''. Specifically, we generate offspring by adding perturbations produced by mirror sampling to the parents: $\bm{\theta}^{i+1}_{j}=\bm{\theta}^{i}_{j_\eta} + \sigma\bm{\epsilon}\ \text{or}\ \textnormal{-}\sigma\bm{\epsilon}$, where $j=0,...,N\textnormal{-}1$, and $\bm{\theta}^{i}_{j_{\eta}}\in\mathcal{P}^i_{\eta}$ is randomly selected with replacement. We can obtain new sub-networks (individuals) by normalizing the connection scores using the sigmoid function and filtering them by the score threshold $s_{th}$. We can repeat steps $\textbf{Evaluation}\textnormal{-}\textbf{Varying}$ for $G\textnormal{-}1$ iterations.

The pseudocode for our GA version is summarized in Algorithm~\ref{ap3}.

\begin{figure}[tbp]
	\centering  
	\includegraphics[width=1.0\columnwidth]{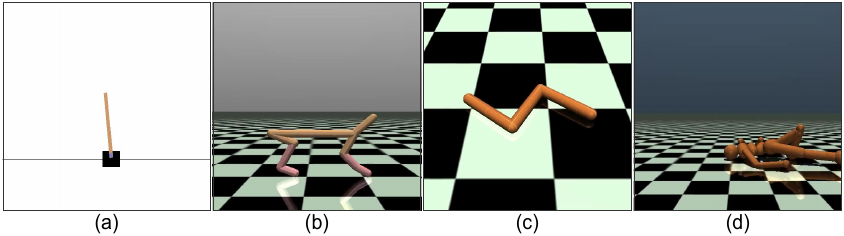}
	\caption{Four robotic control tasks. \textbf{(a) CartPole-v1}: Observation dimension: $n=4$, Action dimension: $m=2$ (discrete), Goal: balance a pole on a cart; \textbf{(b) HalfCheetah-v2}: Observation dimension: $n=17$, Action dimension: $m=6$ (continuous), Goal: make a 2D cheetah robot run as fast as possible; \textbf{(c) Swimmer-v2}: Observation dimension: $n=8$, Action dimension: $m=2$ (continuous), Goal: make a 2D robot swim; \textbf{(d) HumanoidStandup-v2}: Observation dimension: $n=376$, Action dimension: $m=17$ (continuous), Goal: make a 3D two-legged robot standup.}
	\label{fig3}
\end{figure}

\section{Experimental Settings}

\subsection{Evaluation Tasks}

We evaluated our method on a benchmark of robotic control tasks interfaced by OpenAI Gym~\cite{brockman2016openai}, as shown in Fig.~\ref{fig3}. These tasks included classic discrete control tasks like CartPole-v1, as well as more challenging continuous control tasks such as HalfCheetah-v2, Swimmer-v2, and HumanoidStandup-v2, which were run in the fast physics simulator MuJoCo~\cite{todorov2012mujoco}. The original CartPole-v1 task had an upper limit return of $500$, without any manual modification to the environment or reward.

\subsection{Implement Details}

On all tasks, our SPN was a multi-layer perceptron architecture with a 64-unit middle layer, i.e., (sensory layer, middle processing layer, LIF, motor layer, LI), where the number of simulation time step $T'$ was $4$, the decay factor $g$ was $0.75$, the rest and reset potential $v_{rest}$ and $v_{reset}$ were both $0$, and the firing threshold $v_{th}$ was $0.5$. In our GA version, we set the number of generations $G$ to $100$, the number of individuals $N$ in each generation to $200$, the standard deviation $\sigma$ of perturbations to $0.01$, the ratio $\eta$ of truncation selection to $0.25$, and the connection score threshold $s_{th}$ to $0.5$.

For a fair comparison, on all tasks, the DPN was set up with the same number of parameters and architecture as our SPN but with different neuron types, i.e., (sensory layer, middle processing layer, ReLU, motor layer, Tanh). It was optimized in conjunction with a deep value network by PPO~\cite{schulman2017proximal} (one of the most general RL algorithms and the default RL algorithm at OpenAI). The architecture of the deep value network was (sensory layer, middle processing layer, ReLU, output layer), where the output layer with 1 unit estimated the value of the input observation.
The PPO implementation and its hyper-parameter configurations were from OpenAI Spinning Up~\cite{SpinningUp2018}. Our DPN baseline achieved similar or better performance than reported in the literature~\cite{schulman2017proximal,kangin2019policy}. The detailed hyper-parameter configurations and optimization process of PPO were shown in Appendix.


To address concerns regarding reproducibility~\cite{henderson2018deep}, we used the original set of tasks from Gym~\cite{brockman2016openai} without modifying the reward or environment. We conducted $10$ independent runs with different random seeds for all experiments.
In GA, we ran each task for $100$ generations and evaluated the ``elite'' sub-network of each generation with $10$ episodes, reporting the average return over $10$ runs. In PPO, following previous works~\cite{schulman2017proximal,kangin2019policy}, we ran each task for $1e6$ steps and evaluated the final learned DPN with $10$ episodes, reporting the average return over $10$ runs as its performance level.
All experiments were conducted on an AMD EPYC 7742 server.

\section{Results and Discussions}

 Our experiments aimed to accomplish the following objectives: 1) To demonstrate the feasibility of our alternative method, SPN-Connections-GA, for tuning the synaptic connections of SPN by gradient-free GA, {compared to the mainstream DRL method, DPN-Weights-PPO, for tuning the synaptic weights of DPN by gradient-based PPO~\cite{schulman2017proximal}} (Section~\ref{ex1}); 2) To demonstrate the superiority of connection tuning over weight tuning (Section~\ref{ex2}); 3) To demonstrate the necessity of using GA to optimize SPN (Section~\ref{ex3}); 4) To showcase the dynamic evolution process of SPN-Connections-GA (Section~\ref{ex4}).

\subsection{SPN-Connections-GA vs. DPN-Weights-PPO}\label{ex1}
    


In this section, we compare our alternative method, SPN-Connections-GA, with the mainstream DRL method, {DPN-Weights-PPO~\cite{schulman2017proximal}}, in terms of the return gained, data efficiency, and energy efficiency.

\subsubsection{Return Gained}
\begin{figure}[tbp]
	\centering  
	\includegraphics[width=1.0\textwidth]{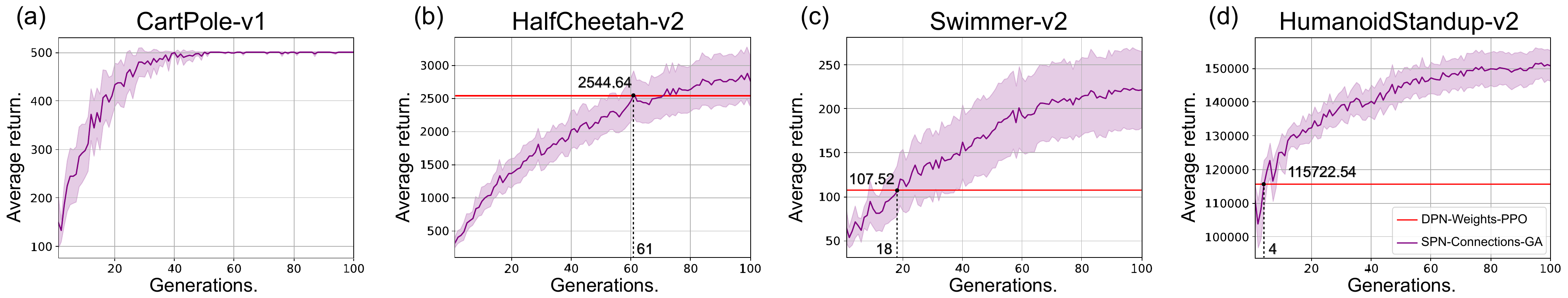}
	\caption{Learning curves for the OpenAI Gym robotic control tasks. The purple areas represented the learning curves of SPN-Connections-GA during 100 generations of evolution, where the solid curves correspond to the mean and the shaded region to half a standard deviation over $10$ runs. The red horizontal line represented the performance level of {DPN-Weights-PPO~\cite{schulman2017proximal}}.}
	\label{fig4}
\end{figure}

Fig.~\ref{fig4} (a) shows that on the classical discrete control task CartPole-v1, SPN-Connections-GA reached the upper limit return of $500$, preliminarily verifying the feasibility of our method. Furthermore, as demonstrated in Fig.~\ref{fig4} (b)--(d), on more challenging MuJoCo continuous control tasks HalfCheetah-v2, Swimmer-v2, and HumanoidStandup-v2, our SPN-Connections-GA was able to achieve the performance level of {DPN-Weights-PPO~\cite{schulman2017proximal}} (the red horizontal line in the Fig.).

\begin{table*}[tbp]
\caption{The exact data efficiency tradeoffs between SPN-Connections-GA and {DPN-Weights-PPO~\cite{schulman2017proximal}}. $\mathcal{R}$ denoted the performance level of {DPN-Weights-PPO~\cite{schulman2017proximal}} after using $1e6$ steps of environment interaction. $\gamma$ denoted a measure of data efficiency, defined in Equation~(\ref{de}).}
\centering
\begin{tabular}{@{}c|cccc@{}}
\hline
Task               & $\mathcal{R}$ &$\mathcal{G}$ &$\mathcal{E}$ & $\gamma$  \\ \hline\hline
HalfCheetah-v2     & $2544.64$ &$61$& $1.83e7$& $\textbf{18.3}$  \\
Swimmer-v2         & $107.52$  &$18$& $5.4e6$& $\textbf{5.4}$   \\
HumanoidStandup-v2 & $115722.54$ & $4$& $1.2e6$&$\textbf{1.2}$    \\ \hline
\end{tabular}
\label{ratio}
\end{table*}


\subsubsection{Data Efficiency}
Although the data efficiency of GA is typically low, we found that the data efficiency $\gamma$ of our SPN-Connections-GA was surprisingly good. In our approach, each data (step) that interacted with the environment was used only once. As Table~\ref{ratio} shows, using no more than $18.3$x as much data (i.e., $18.3$x on HalfCheetah-v2, $5.4$x on Swimmer-v2, and $1.2$x on HumanoidStandup-v2), our SPN-Connections-GA was able to achieve the same performance level as {DPN-Weights-PPO~\cite{schulman2017proximal}}, which used $1e6$ steps of environment interaction. However, {DPN-Weights-PPO~\cite{schulman2017proximal}} was more data-efficient because it maintained a vast replay buffer for storing data interacting with the environment, which could be utilized multiple times during optimization. In contrast, our SPN-Connections-GA did not use data repeatedly, did not have a deep value network, and did not perform BP during optimization, resulting in much lower energy consumption, which offset the slight decrease in data efficiency. We defined the specific data efficiency calculation method as follows.

For each MuJoCo control task, we first calculate the steps $\mathcal{E}$ of environment interaction required for SPN-Connections-GA to achieve the performance level $\mathcal{R}$ of {DPN-Weights-PPO~\cite{schulman2017proximal}}. Then, we calculate the ratio $\gamma$ of $\mathcal{E}$ to standard PPO steps ($1e6$) as a measure of data efficiency.
\begin{equation}
\begin{aligned}
    \mathcal{A} &= N\times T + 10\times 10 \times T  \\
    \mathcal{E} &= \mathcal{G} \times \mathcal{A}\\
    \gamma & = \mathcal{E}/{1e6} \\
\end{aligned}
\label{de}
\end{equation}
where $\mathcal{G}$ was the number of generations required to achieve the performance level of {DPN-Weights-PPO~\cite{schulman2017proximal}} (inferred from the vertical dashed line in Fig.~\ref{fig4} (b) -- (d)), $\mathcal{A}$ denoted the number of interaction steps consumed per generation of optimization, $T$ was the length of the episode ($T$=$1000$ in all the tasks we used), and $N$=$200$ was the number of individuals in each generation (each individual needed to interact with the environment to get an episode and use the return of the episode as its fitness score). In each generation, we needed to evaluate each of the top 10 individuals in 10 additional episodes to obtain the ``elite'' individual. Thus, $\mathcal{A}$ was $3e5$. The exact data efficiency tradeoffs were listed in Table~\ref{ratio}.



\subsubsection{Energy Efficiency}
In this section, we analyze the energy efficiency of our SPN-Connections-GA compared to {DPN-Weights-PPO~\cite{schulman2017proximal}}, considering both network inference and optimization. As shown in Table~\ref{inference}, our SPN's network inference energy consumption is 1.3x, 1.2x, and 1.1x lower on HalfCheetah-v2, Swimmer-v2, and HumanoidStandup-v2, respectively, compared to DPN. For optimization, as shown in Table~\ref{optim}, our GA's energy consumption is 6.5x, 21.9x, and 87.9x lower on HalfCheetah-v2, Swimmer-v2, and HumanoidStandup-v2, respectively, compared to PPO. The specific energy consumption calculation method was defined in the next section.

\begin{table}[tbp]
\caption{The network inference energy consumption comparison of our SPN and DPN. The input layer in our SPN was non-spiking, so it’s energy was same as DPN. Unit: pJ.}
\centering
\begin{tabular}{@{}c|ccc@{}}
\hline
Task               & $E^{infer}_{DPN}$  &$E^{infer}_{SPN}$   &$\frac{E^{infer}_{DPN}}{E^{infer}_{SPN}}$   \\ \hline\hline
HalfCheetah-v2     & $6.77e3$ &$5.2e3$& $\textbf{1.3}$ \\
Swimmer-v2         & $2.94e3$  &$2.41e3$& $\textbf{1.2}$   \\
HumanoidStandup-v2 & $1.16e5$ & $1.1e5$& $\textbf{1.1}$   \\\hline
\end{tabular}
\label{inference}
\end{table}

\begin{table}[tbp]
\caption{The optimization energy consumption comparison of our GA and PPO. Unit: pJ}
\centering
\begin{tabular}{@{}c|ccc@{}}
\hline
Task               & $E^{optim}_{PPO}$  &$E^{optim}_{GA}$   &$\frac{E^{optim}_{PPO}}{E^{optim}_{GA}}$   \\\hline\hline
HalfCheetah-v2     & $6.16e11$ &$9.52e10$& $\textbf{6.5}$ \\
Swimmer-v2         & $2.85e11$  &$1.3e10$& $\textbf{21.9}$   \\
HumanoidStandup-v2 & $1.16e13$ & $1.32e11$& $\textbf{87.9}$   \\\hline
\end{tabular}
\label{optim}
\end{table}


\subsubsection{{Theoretical Analysis of Energy Consumption}}


{For inference, we first analyzed the computation patterns of DPN and SPN. For DPN, each operation involved a dot-product that required one floating-point multiplication and one accumulation (MAC). In contrast, for SPN, each operation involved only an accumulation (AC) due to the use of binary spikes (0 or 1). Following~\cite{rathi2021diet}, we computed the inference energy consumption for DPN and SPN using 45nm CMOS technology. Previous literature has shown that the energy consumption for a 32-bit DNN MAC operation ($e_{MAC}$=$4.6 pJ$) is 5.1 times more than that of an SNN AC operation ($e_{AC}$=$0.9 pJ$)\cite{horowitz20141}.}

{Following~\cite{rathi2021diet}, we define the theoretical energy consumption for network inference as the product of the number of operations in each layer of the network and the energy required for a single operation, formulated as follows:
\begin{equation}
\begin{aligned}
    E^{infer}_{DPN} &= \sum\nolimits_{l} \#OP_{DPN} \times e_{MAC} \\
    E^{infer}_{SPN} &= \sum\nolimits_{l} \#OP_{SPN} \times e_{AC}
\end{aligned}
\end{equation}
}

{Following~\cite{rathi2021diet}, in DPN, we defined the number of operations at layer $l$ as:
\begin{equation}
    \#OP_{DPN} =  f_{in} \times f_{out} 
\end{equation}
where $f_{in}(f_{out})$ was the dimension of input (output) features at layer $l$. The number of operations at layer $l$ in SPN was defined as:
\begin{equation}\label{eq:operations_snn}
    \begin{aligned}
    \#OP_{SPN} &= rate_l \times \#OP_{DPN} \\
    rate_l &= \frac{\# totalSpikes_l}{\#n_l}
    \end{aligned}
\end{equation}
where $rate_l$ was the total spikes at layer $l$ over time window averaged over the number of neurons $n_l$ at layer $l$ (averaged over all interaction steps).}

{We define the theoretical energy consumption of network optimization as the theoretical energy consumption of network inference multiplied by the number of network inferences during the optimization process:
\begin{equation}
\label{optimization}
\begin{aligned}
\begin{split}
E^{optim}_{PPO} &= E^{infer}_{DPN} \times (N^{forward}_{DPN}+N^{backward}_{DPN}) \\
    &+ E^{infer}_{DVN} \times (N^{forward}_{DVN}+N^{backward}_{DVN}) \\
    E^{optim}_{GA} &= E^{infer}_{SPN} \times N^{forward}_{SPN} 
\end{split}
\end{aligned}
\end{equation}
}

\begin{table}[tbp]
\centering
\caption{The specific forward and back propagation times of data in the network.}
\begin{tabular}{@{}cc|cc|cc@{}}
\hline
                    &                    & \multicolumn{2}{c|}{Policy Network} & \multicolumn{2}{c}{Value Network} \\ 
                    &                    & forward          & backward         & forward         & backward        \\\hline\hline
\multicolumn{2}{c|}{PPO}                 &     2.6e7             &     2.5e7             &      2.6e7           &          2.5e7       \\ \hline
\multirow{3}{*}{GA} & HalfCheetah     &    1.83e7              &       --           &         --        &          --       \\
                    & Swimmer        &      5.4e6            &     --             &     --            &--                 \\
                    & HumanoidStandup &         1.2e6         &         --         &   --              & --                 \\\hline
\end{tabular}
\label{number}
\end{table}


{Here, we calculate the number of network inferences during the optimization process. First, the PPO optimization process includes both the forward inference and backpropagation of the DPN, as well as the forward inference and backpropagation of the deep value network (DVN). To simplify the energy consumption calculations, we consider both the network backpropagation and forward inference to have the same energy consumption. Specifically, for each task, the PPO optimization requires interaction with the environment to collect a total of $1e6$ data (steps). Each time a piece of data is collected, both the DPN (inferred an action based on observation) and the DVN (estimated the value of the observation) need to conduct a forward inference. For every $1e3$ data collected, $25$ epochs of optimization (including forward inference and backpropagation of DPN and DVN) are conducted based on these data. During GA optimization, there is only forward inference of SPN, and the number of forward inferences corresponds to the number of data ($N^{forward}_{SPN}$ in Equation~(\ref{optimization}) equals $\mathcal{E}$ in Equation~(\ref{de})). The forward and backpropagation times of data in the network are listed in Table~\ref{number}. It's worth noting that we calculated the optimization energy consumption of our SPN-Connections-GA when achieving the same performance level as {DPN-Weights-PPO~\cite{schulman2017proximal}}.}

\subsection{SPN-Connections-GA vs. SPN-Weights-GA}\label{ex2}
 In this section, we verified the superiority of connection tuning over weight tuning by comparing the performance of SPN-Connections-GA with SPN-Weights-GA.

\begin{figure}[tbp]
	\centering  
	\includegraphics[width=1.0\textwidth]{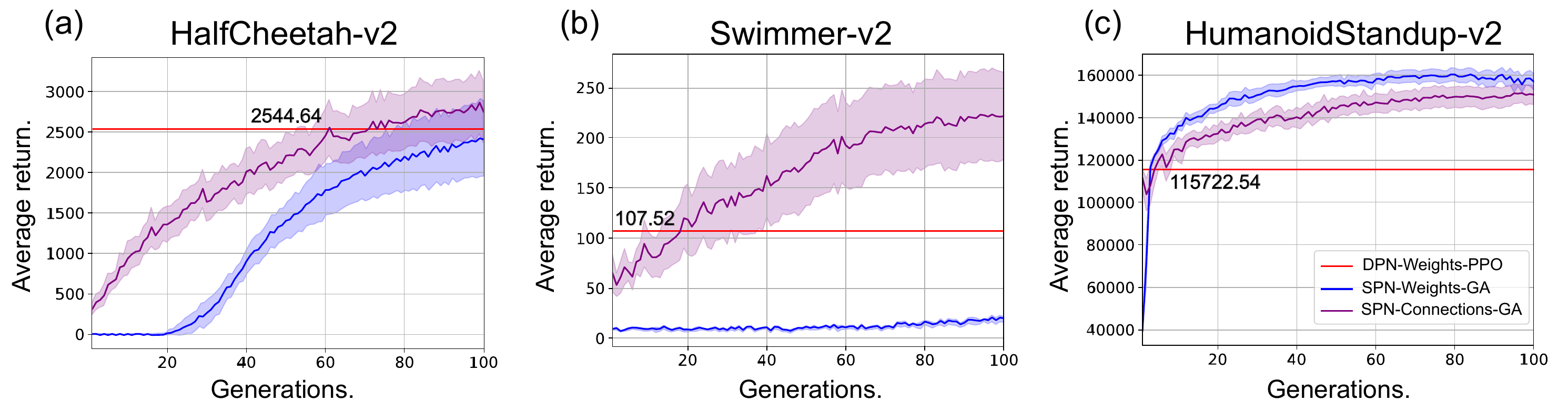}
	\caption{Learning curves for the MuJoCo continuous control tasks. The purple areas represented the learning curves of SPN-Connections-GA during 100 generations of evolution, where the solid curves correspond to the mean and the shaded region to half a standard deviation over $10$ runs. The blue areas represented the learning curves of SPN-Weights-GA during 100 generations of evolution, where the solid curves correspond to the mean and the shaded region to half a standard deviation over $10$ runs. The red horizontal line represented the performance level of {DPN-Weights-PPO~\cite{schulman2017proximal}}.}
	\label{fig5}
\end{figure}


Fig.~\ref{fig5} (a)--(c) shows that SPN-Connections-GA achieved a higher average return on HalfCheetah-v2 and Swimmer-v2 compared to SPN-Weights-GA. Additionally, SPN-Connections-GA achieved a comparable average return on HumanoidStandup-v2. It's worth noting that even after 100 generations of evolution, SPN-Weights-GA could not achieve the same performance level as {DPN-Weights-PPO~\cite{schulman2017proximal}} on HalfCheetah-v2 and Swimmer-v2. In fact, SPN-Weights-GA struggled to learn and was only able to achieve a shallow level of performance on Swimmer-v2. These results suggest that an "elite" sub-network of SPN with random weights can achieve a comparable or better return than the full network with learned weights tuned by the same GA.

\subsection{SPN-Connections-GA vs. SPN-Connections-PPO}\label{ex3}

\begin{figure}[tbp]
	\centering  
	\includegraphics[width=1.0\textwidth]{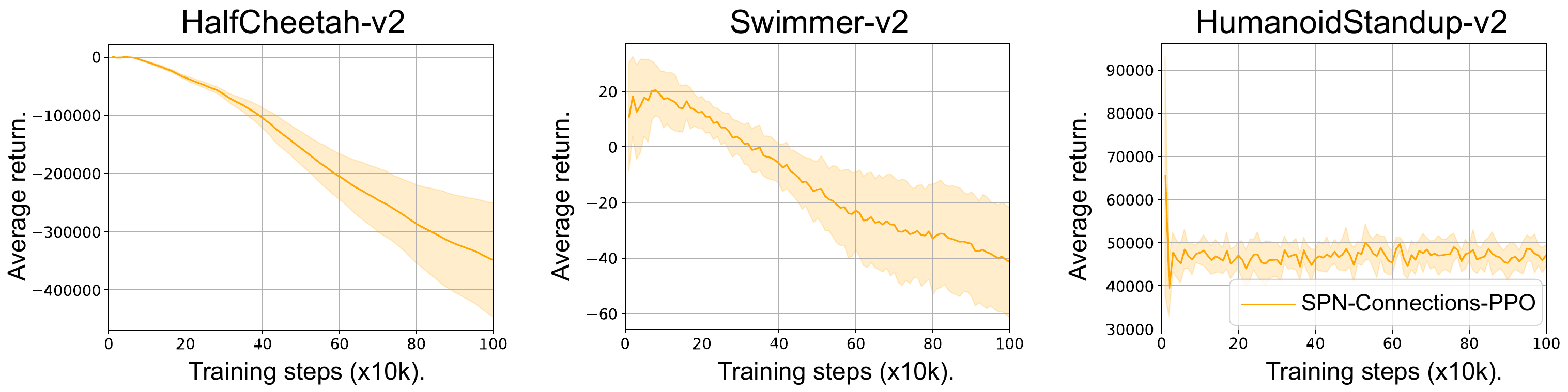}
	\caption{Learning curves for the MuJoCo continuous control tasks. The yellow areas represented the learning curves of SPN-Connections-PPO, where the solid curves correspond to the mean and the shaded region to half a standard deviation over $10$ runs.}
	\label{fig6}
\end{figure}

In this section, we aimed to verify the necessity of using GA to optimize SPN by comparing the performance of SPN-Connections-GA with SPN-Connections-PPO. We observed that directly replacing DPN with our SPN and integrating it into the gradient-based PPO optimization framework did not produce satisfactory results. As shown in Fig.~\ref{fig6}, the learning curve of SPN-Connections-PPO did not converge, and its average return even decreased over time during training. On the other hand, as demonstrated in Fig.~\ref{fig4}, SPN-Connections-GA performed well on the same tasks. The suboptimal performance of SPN-Connections-PPO might be attributed to the fact that our SPN, which relied on discrete spike communication, was a low-precision network. As a result, using a gradient-based PPO algorithm to optimize such a network could produce biased low-precision gradient estimates, leading to suboptimal performance. In contrast, black-box optimization algorithms like GA might be better suited for optimizing low-precision networks such as SPN.

\subsection{The evolution process of SPN-Connections-GA}\label{ex4}

For MoJoCo continuous control tasks HalfCheetah-v2 and Swimmer-v2, we randomly sampled one of 10 independent evolutionary runs and visualized the dynamic evolution process of the sub-network of SPN. The video URLs of dynamic evolution were listed in Table~\ref{video}.

\begin{table}[htbp]
\centering
\caption{The video URLs of dynamic evolution.}
\begin{tabular}{@{}cc@{}}
\toprule
Task & URL \\ \midrule\midrule
HalfCheetah-v2  &   \url{https://www.youtube.com/watch?v=_o0v3LtmU-U}          \\ 
Swimmer-v2  &   \url{https://www.youtube.com/watch?v=JmxR4deP4Cc}  \\\bottomrule
\end{tabular}
\label{video}
\end{table}

\section{Conclusion}


In this paper, we propose a novel approach for optimizing a biologically-inspired SPN using a gradient-free GA as an energy-efficient alternative to DRL. Unlike previous methods that primarily focused on tuning synaptic weights, we adopt a new perspective of optimizing synaptic connections to evolve ``elite'' sub-networks of the SPN capable of solving complex tasks. Our extensive experiments on several challenging robotic control tasks demonstrate that our approach can achieve the performance level of mainstream DRL methods while significantly reducing energy consumption. Moreover, our analyses confirm the superiority of connection tuning over weight tuning and the necessity of using GA to optimize SPN.

\section{Acknowledgments}
This work was supported by the Beijing Nova Program (Grant No. 20230484369), the Strategic Priority Research Program of Chinese Academy of Sciences (Grant No. XDA27010404), the Shanghai Municipal Science and Technology Major Project (Grant No. 2021SHZDZX), and the Youth Innovation Promotion Association of the Chinese Academy of Sciences.


\bibliographystyle{bst/sn-mathphys}
\bibliography{sn-bibliography}

\newpage

\section*{Appendix}

\subsection*{The Hyper-parameter Configurations and Optimization Process of PPO}\label{ap1}

For each MuJoCo continuous control task, we ran $1e3$ iterations, $1e3$ steps per iteration, and a total of $1e6$ steps of environment interaction. The deep policy network was trained in $25$ epochs per iteration with a mini-batch size of $100$, using Adam optimizer with a learning rate of $1e\textnormal{-}5$. The training configurations of the deep value network were the same as the deep policy network, but the learning rate was $1e\textnormal{-}4$. The discount factor $\gamma$ for reward was $0.99$, and the discount factor $\lambda$ used in generalized advantage estimation was $0.95$. The clip ratio $\epsilon_{\text{clip}}$ for PPO was $0.2$.  We presented the optimization process of PPO in Algorithm~\ref{alg:ppo}.

\begin{algorithm*}[htbp]
\caption{The optimization process of PPO.}
\label{alg:ppo}

Initialize deep policy network $\pi(\bm{a}\vert\bm{o})$ and deep value network $V(\bm{o})$ with parameter $\bm{\phi}_\pi$ and $\bm{\phi}_V$.

\FOR{$iter=1$ to $1e3$}

\ \ \ \ Run $\pi_{\phi_\pi}$ to collect a set of episodes $\mathcal{D}=\{\epsilon^k\}$ containing $\vert\mathcal{D}\vert$ episodes, each $\epsilon^k$ is a episode contain $\vert\epsilon^k\vert$ data (samples or steps), $\epsilon^{k} := \{(\bm{o}^{k,t}, \bm{a}^{k,t}, r^{k,t+1}, \bm{o}^{k, t+1})\}$, $t \in [\vert\epsilon^k\vert]$.

\ \ \ \ \ \ \ \ Compute cumulative reward $R^{k,t}$ for each step $t$ in every episode $k$ using discount factor $\gamma$.

\ \ \ \ \ \ \ \ Update deep value network by minimizing the mean-square error:

\ \ \ \ \ \ \ \ $\bm{\phi}_V \leftarrow \mathop{\arg\min}\limits_{\bm{\phi}_V} \frac{1}{\sum\limits_{k} \vert\epsilon^k\vert} \sum\limits_{\epsilon^k \in D} \sum\limits_{t=0}^{\vert\epsilon^k\vert} \left (V(\bm{o}^{k,t}) - {R}^{k,t} \right )^2$

\ \ \ \ \ \ \ Estimate advantage ${A}^{k,t}$ for each step $t$ in every episode $k$ using generalized advantage estimation and deep value network $V_{\bm{\phi}_V}(o)$.

\ \ \ \ \ \ \ Update the deep policy network by minimizing the following objective (the minimization is solved using Adam optimizer):

\ \ \ \ \ \ \ \[
\bm{\phi}_\pi \leftarrow \mathop{\arg\min}\limits_{\bm{\phi}_\pi^\prime} \frac{1}{\sum\limits_k\vert\epsilon^k\vert} \left [ \sum\limits_{\epsilon^k \in D} \sum\limits_{t=0}^{\vert\epsilon^k\vert} \min \left ( r_{\bm{\phi}_\pi^\prime}(\bm{a}^{k,t} \vert \bm{o}^{k,t}) {A}^{k,t}, g(r_{\bm{\phi}_\pi^\prime}(\bm{a}^{k,t} \vert \bm{o}^{k,t})) {A}^{k,t} \right ) \right ]
\]

\ \ \ \ \ \ \ where $r_{\bm{\phi}_\pi^\prime}(\bm{a}^{k,t} \vert \bm{o}^{k,t}) := \frac{\pi_{\bm{\phi}_\pi^\prime}(\bm{a}^{k,t} \vert\bm{o}^{k,t})}{\pi_{\bm{\phi}_\pi} (\bm{a}^{k,t} \vert\bm{o}^{k,t})}$, $g(r):=\mathrm{clip}(r_{\bm{\phi}_\pi^\prime}(\bm{a}^{k,t} \vert \bm{o}^{k,t}), 1 - \epsilon_\text{clip}, 1 + \epsilon_\text{clip})$.

{\bfseries Return:} Final learned deep policy network.
\end{algorithm*}

\newpage

\begin{figure}[H]%
\centering
\includegraphics[width=0.3\textwidth]{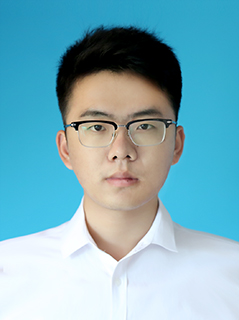}
\end{figure}

\noindent{\bf Duzhen Zhang}\quad received the B.Sc. degree in software engineering from Shandong University, China in 2019. He is a Ph.D. candidate in both the Institute of Automation Chinese Academy of Sciences and the University of Chinese Academy of Sciences. His current interests include theoretical research on reinforcement learning, natural language processing, and Spiking Neural Networks.

E-mail: zhangduzhen2019@ia.ac.cn
ORCID iD: 0000-0002-4280-431X

\begin{figure}[H]%
\centering
\includegraphics[width=0.3\textwidth]{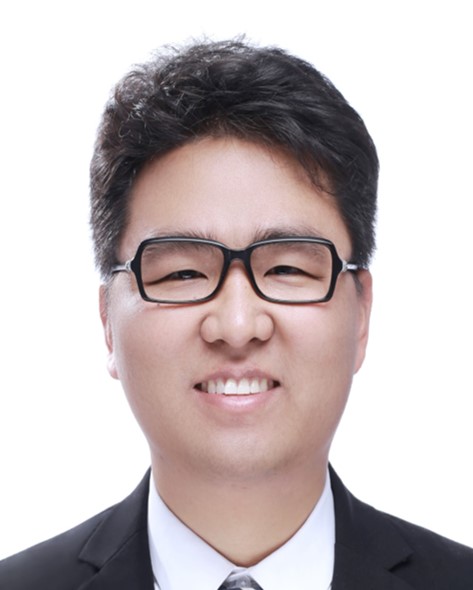}
\end{figure}

\noindent{\bf Tielin Zhang}\quad received the Ph.D. degree from the Institute of Automation Chinese Academy of Sciences, Beijing, China, in 2016. He is an Associate Professor at the Research Center for Brain-Inspired Intelligence, Institute of Automation, Chinese Academy of Sciences, Beijing, China. His current interests include theoretical research on neural dynamics and Spiking Neural Networks (more information is in https://bii.ia.ac.cn/$\sim$tielin.zhang/).

E-mail: tielin.zhang@ia.ac.cn (Corresponding author)
ORCID iD: 0000-0002-5111-9891

\begin{figure}[H]%
\centering
\includegraphics[width=0.3\textwidth]{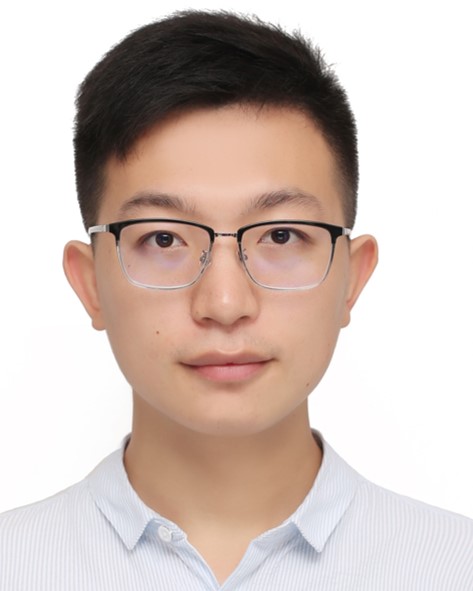}
\end{figure}

\noindent{\bf Shuncheng Jia}\quad is a Ph.D. candidate in both the Institute of Automation Chinese Academy of Sciences and the University of Chinese Academy of Sciences. His current interests include theoretical research on neural dynamics, auditory signal processing, and Spiking Neural Networks.

E-mail: jiashuncheng2020@ia.ac.cn

\begin{figure}[H]%
\centering
\includegraphics[width=0.3\textwidth]{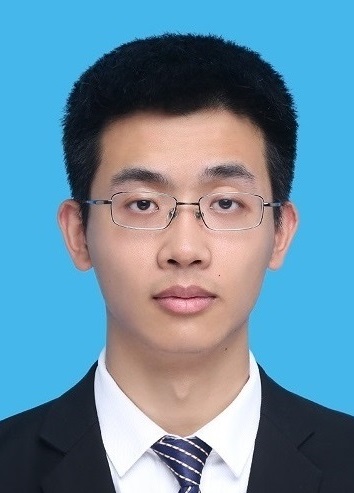}
\end{figure}

\noindent{\bf Qingyu Wang}\quad is an M.D. candidate in both the Institute of Automation Chinese Academy of Sciences and the University of Chinese Academy of Sciences. His current interests include theoretical research on neural dynamics and Spiking Neural Networks.

E-mail: wangqingyu2022@ia.ac.cn

\begin{figure}[H]%
\centering
\includegraphics[width=0.3\textwidth]{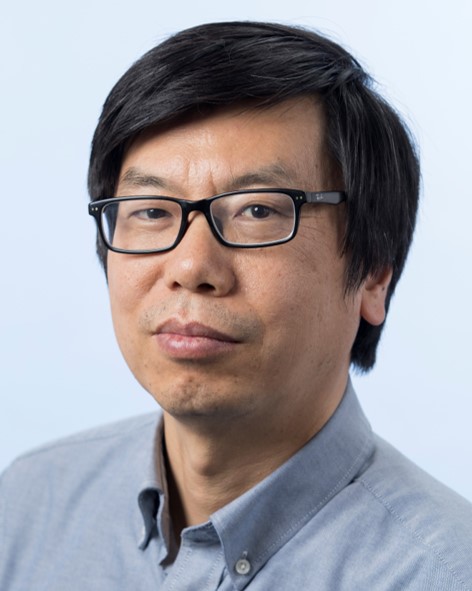}
\end{figure}

\noindent{\bf Bo Xu}\quad is a professor, the director of the Institute of Automation Chinese Academy of Sciences, and also deputy director of the Center for Excellence in Brain Science and Intelligence Technology, Chinese Academy of Sciences. His main research interests include brain-inspired intelligence, brain-inspired cognitive models, natural language processing and understanding, brain-inspired robotics.

E-mail: xubo@ia.ac.cn (Corresponding author)
ORCID iD: 0000-0002-1111-1529 

\end{document}